\documentclass[lettersize,journal]{IEEEtran}
\usepackage{amsmath,amsfonts}
\usepackage{algorithmic}
\usepackage{algorithm}
\usepackage{array}
\usepackage[caption=false,font=normalsize,labelfont=sf,textfont=sf]{subfig}
\usepackage{textcomp}
\usepackage{stfloats}
\usepackage{url}
\usepackage{verbatim}
\usepackage{graphicx}
\usepackage{cite}
\usepackage{booktabs}
\usepackage{amssymb}
\usepackage{pifont}
\usepackage{multirow}
\usepackage{hyperref}
\usepackage[table]{xcolor}

\begin{document}

\title{Rethinking Token-wise Feature Caching: \\ Accelerating Diffusion Transformers with Dual Feature Caching}

\author{
\textbf{Chang Zou\textsuperscript{\rm 1,*}},
\textbf{Shikang Zheng\textsuperscript{\rm 1,2,*}},
\textbf{Evelyn Zhang\textsuperscript{\rm 1}},
\textbf{Runlin Guo\textsuperscript{\rm 3}},
\textbf{Haohang Xu\textsuperscript{\rm 4}},
\textbf{Zhengyi Shi\textsuperscript{\rm 1}}\\
\textbf{Conghui He\textsuperscript{\rm 5}},
\textbf{Xuming Hu\textsuperscript{\rm 6}},
\textbf{Linfeng Zhang\textsuperscript{\rm 1,\dag}}\\
\textsuperscript{\rm 1} Shanghai Jiao Tong University
\textsuperscript{\rm 2} South China University of Technology
\textsuperscript{\rm 3} Beihang University\\
\textsuperscript{\rm 4} Huawei Technologies Ltd
\textsuperscript{\rm 5} Shanghai AI Lab
\textsuperscript{\rm 6} Hong Kong University of Science and Technology (Guangzhou)
\thanks{
* Equal contribution.\\
${}^\dag$ Denotes the corresponding author.
}
}

\maketitle

\section{Abstract}
Diffusion Transformers (DiT) have become the dominant methods in image and video generation yet still suffer substantial computational costs. 
As an effective approach for DiT acceleration, feature caching methods are designed to cache the features of DiT in previous timesteps and reuse them in the next timesteps, allowing us to skip the computation in the next timesteps. Among them, token-wise feature caching has been introduced to perform different caching ratios for different tokens in DiTs, aiming to skip the computation for unimportant tokens while still computing the important ones. 
In this paper, we propose to carefully check the effectiveness in token-wise feature caching with the following two questions: (1) Is it really necessary to compute the so-called "important" tokens in each step? (2) Are so-called important tokens really important?
Surprisingly, this paper gives some counter-intuition answers, demonstrating that 
consistently computing the selected ``important tokens'' in all steps is not necessary. The selection of the so-called ``important tokens'' is often ineffective, and even sometimes shows inferior performance than random selection. 
Based on these observations, this paper introduces dual feature caching referred to as DuCa, which performs aggressive caching strategy and conservative caching strategy iteratively and selects the tokens for computing randomly. Extensive experimental results demonstrate the effectiveness of our method in DiT, PixArt, FLUX, and OpenSora, demonstrating significant improvements than the previous token-wise feature caching.



\begin{IEEEkeywords}
Diffusion Acceleration, Feature Caching.
\end{IEEEkeywords}

\section{Introduction}
\label{sec:intro}

Diffusion Models (DMs)\cite{ho2020DDPM} have shown remarkable success in both image \cite{rombach2022SD} and video generation \cite{blattmann2023SVD}. Traditional DMs often use U-Net denoising structures, such as Stable Diffusion (SD)\cite{rombach2022SD} and Stable Video Diffusion (SVD)\cite{blattmann2023SVD}, which have already demonstrated impressive results. Most recently, with the introduction of Diffusion Transformers (DiT)\cite{peebles2023dit}, the quality of visual generation has significantly improved further, propelling the development of numerous downstream applications \cite{Feng2024DiT4EditDT}.  However, the quality improvement is also accompanied by a significant increase in computational demands, reducing the inference efficiency of transformers and making their deployment in practical scenarios a pressing challenge. 

\begin{figure}
    \centering
    \includegraphics[width=\linewidth]{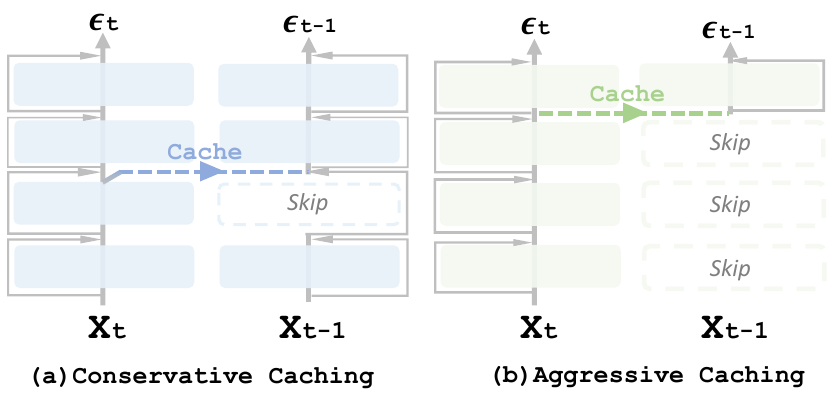}
    \caption{Comparison between conservative caching (a) and aggressive caching (b): Conservative caching skips some tokens instead of all tokens in each layer to maintain the generation quality, but also leads to lower acceleration ratios. $t$ and $t-1$ denote the refreshing and caching timesteps, respectively.}
    \label{fig:aggressive-conservative}
    \vspace{-0.5cm}
\end{figure}

Recently, feature caching has become one of the most popular techniques for diffusion acceleration. Motivated by the high similarity between features in the adjacent timesteps, \emph{aggressive} feature caching caches the features computed in a \emph{freshing timestep}, and then reuse them in the following several \emph{caching steps}, as shown in Figure~\ref{fig:aggressive-conservative}(b). Such a fresh-then-cache strategy enables the diffusion model to save computational costs in the caching step and thus leads to significant acceleration. 
However, since the similarity of features in non-adjacent timesteps decreases quickly, the error introduced by aggressive feature caching can be accumulated at an exponential speed, leading to a significant drop in generation quality, and making aggressive feature caching impractical.
 
To solve this problem, recent work \cite{chen2024delta-dit, ma2024l2c, selvaraju2024fora, zou2024accelerating} introduces some computations during the caching steps in certain layers and tokens, which makes the cached features in previous steps corrected for the current caching step. Among them, token-wise feature caching (\texttt{ToCa}) has been proposed to selectively compute the important tokens while skipping the unimportant tokens with the cache, reducing the accumulated error from feature caching, but also making it difficult to achieve a higher acceleration ratio. In this paper, we rethink and distill \texttt{ToCa}. Specifically, we focus on the following two questions. 

\begin{figure}
    \centering
    \includegraphics[width=1.0\linewidth]{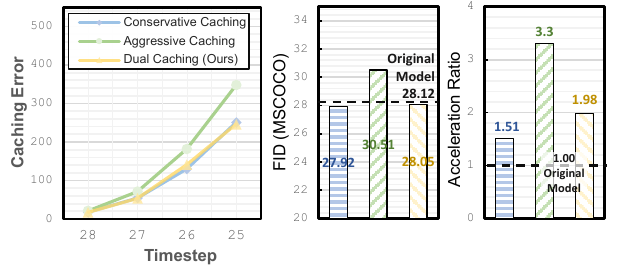}
    \vspace{-5mm}
    \caption{ The caching error and acceleration ratio of three feature caching strategies.
    The caching error is defined as the $L2$-norm distance between $x_t$ computed with and without feature caching. 
    }
    \label{fig:aggressive-conservative-loss}
    \vspace{-4mm}
\end{figure}

\noindent\emph{\textbf{(I) Is it necessary to compute the important tokens in all caching steps?}}
\texttt{ToCa} computes the important tokens in each caching step to reduce the accumulated error from feature caching.  Figure~\ref{fig:aggressive-conservative-loss} shows the caching error from the aggressive (marked in green) and conservative caching (\texttt{ToCa}, marked in blue).
It is observed that the caching error from \texttt{ToCa} increases slowly as the caching step grows, while the error in the aggressive caching method accumulates rapidly with each caching step, which verifies the effectiveness of \texttt{ToCa}.
However, we find that the caching error of the aggressive and conservative caching exhibit a close value at the first caching step (\emph{i.e.}, timestep 27 in Figure~\ref{fig:aggressive-conservative-loss}), suggesting that computing the important tokens in \texttt{ToCa} is not always necessary. Concretely, when the error from caching has not been accumulated yet, computing important tokens does not bring benefits and can be removed for better efficiency.

\noindent \textbf{Solution: Dual Feature Caching}: Based on this observation, we propose \texttt{DU}al \texttt{CA}ching, which leverages the advantages of aggressive and conservative caching methods. In each caching cycle, \texttt{DuCa} initializes the cache by performing the full computation in the \emph{fresh} timestep. Then, it performs the one-step aggressive caching for high-ratio acceleration, followed by one-step conservative caching to fix the cache error. This alternative caching strategy can be repeated multiple times to achieve the best balance between efficiency and quality.

\noindent\textbf{\emph{(II) Are so-called important tokens really important?}} The basic assumption of \texttt{ToCa} is that there exist some tokens that are sensitive or robust to feature caching. Based on this motivation, \texttt{ToCa} introduces four kinds of scores to describe the importance of each token based on their attention scores, spatial position, and frequency of being cached, respectively. 
Unfortunately, these scores not only introduce additional computation costs, but also lead to the incompatibility of FlashAttention, one of the default efficient operators in hardware. Hence, it results in around 100\% increments in latency and improves the memory overhead from $O(N)$ to $O(N^2)$.
To solve this problem, we carefully study the effectiveness of each score in \texttt{ToCa}.
Surprisingly, as shown in Table~\ref{table:token-selection} and will be discussed later, we find that selecting tokens with the highest attention scores in \texttt{ToCa} sometimes shows inferior performance than random token selection, indicating previous methods may misunderstand the importance of tokens.  

\noindent \textbf{Solution: Selecting tokens with low similarity}: After carefully trying all kinds of importance scores for token selection, we find that choosing a subset of tokens that has the smallest similarity to other tokens is the only solution to beat random token selection. On the other hand, choosing a subset of tokens with maximal similarity to each other almost achieves the worst performance that we can have. Based on this observation, we argue that \emph{token selection should focus on their duplication instead of their importance}. Thus, random selection can be a good choice since randomness guarantees choosing tokens with different semantic information and leads to no additional computation or memory overhead.


Extensive experiments have demonstrated the performance of our method in abundant benchmarks, including text-to-image generation with FLUX.1-dev, PixArt-$\alpha$, text-to-video generation with OpenSora, and class-conditioned image generation with DiT. For instance,  $3.45\times$ acceleration on FLUX.1-dev and $2.50\times$ acceleration on OpenSora with almost lossless quality. In summary, our main contributions are as follows:

\begin{enumerate}

    \item We rethink the necessity of computing ``important'' tokens in token-wise feature caching, and introduce \texttt{DuCa}, a dual feature caching strategy to perform the two caching methods alternatively, maintaining the generation quality while achieving a higher acceleration ratio.

    \item We rethink the tokens selection methods in token-wise feature caching, finding that previous selection methods show inferior performance than random token selection, and explaining this phenomenon from the perspective of duplication. By just replacing previous methods with random selection, we achieve good compatibility with FlashAttention, leading to $\sim2\times$ speed up in latency.
    
    \item Extensive experiments have demonstrated \texttt{DuCa} in conditional image generation, text-to-image generation, and text-to-video generation, outperforming previous methods by a large margin. 
    Our codes have been released.
    
\end{enumerate}



\section{Related Works}
\label{sec:Related Works}
\noindent\textbf{Acceleration of Diffusion Models}
To address the high computational costs commonly associated with DMs including DiT, various acceleration methods have been proposed. Existing acceleration methods can generally be divided into two categories: Designing sampling methods to reduce the number of steps needed to transition from noise to high-quality images, and directly accelerating the denoising model.
Sampling-based acceleration methods enhance the efficiency of DMs by simply reducing the sampling steps.
DDIM \cite{songDDIM} leverages a deterministic, non-Markovian reverse process to significantly reduce generation steps while maintaining high-quality outputs. DPM Solver \cite{lu2022dpm}, through analytic derivation, optimizes the reverse process, enabling efficient sampling with fewer steps, while DPM Solver++ \cite{lu2022dpm++} further refines this approach by incorporating noise control and diverse sampling techniques to improve quality and adaptability. Rectified Flow \cite{refitiedflow} employs a recurrent flow framework, utilizing specially designed flow fields to accelerate the diffusion process.

\noindent\textbf{Feature Caching~}Due to the high computational cost of retraining the diffusion models, or even small learnable modules introduce significant costs, limiting model reusability across different scales, resolutions, and tasks, driving interest in training-free acceleration methods.
Faster Diffusion  \cite{rombach2022SD} improves computational efficiency by caching the output of the encoder module on certain steps, while DeepCache  \cite{ma2024deepcache} reduces redundant computations by reusing low-resolution feature information in the skip connections of the U-Net structure. Unfortunately, although both are training-free methods, they are specialized feature caching techniques designed specifically for U-Net-based models like Stable Diffusion \cite{rombach2022SD}, making them difficult to directly adapt to transformer-based models like DiT \cite{chen2024delta-dit}.
Recently, to address the lack of feature reuse solutions for DiT, $\Delta$-DiT \cite{chen2024delta-dit} and Pyramid Attention Broadcast (PAB) \cite{zhao2024PAB} have been introduced. The former focuses on constructing residuals of DiT outputs across different layers, while the latter emphasizes the importance of various types of attention within DiT Blocks containing multiple attention heads. Learning-to-Cache \cite{ma2024l2c} achieves higher acceleration ratios by employing a learnable router to determine whether computations are needed at each layer, but it incurs prohibitive computational costs. These feature caching methods generally lack attention to finer-grained features. \texttt{ToCa} \cite{zou2024accelerating} proposes a method that selects important tokens at both layer and token granularity, allocating more computation to these tokens while caching less important ones, advancing temporal redundancy management to the token level.

\section{Method}

\subsection{Preliminary}

\paragraph{Diffusion Models} Diffusion models \cite{ho2020DDPM} are designed with two main processes: a forward process, where Gaussian noise is incrementally added to a clean image, and a reverse process, where a standard Gaussian noise is gradually denoised to reconstruct the original image. Letting $t$ represent the timestep and $\beta_t$ the noise variance schedule, the conditional probability in the reverse (denoising) process, $p_\theta(x_{t-1} \mid x_t)$, can be modeled as:
\begin{equation}
\label{eq: new_reverse_process}
    \mathcal{N} \left( x_{t-1}; \frac{1}{\sqrt{\alpha_t}} \left( x_t - \frac{1 - \alpha_t}{\sqrt{1 - \bar{\alpha}_t}} \epsilon_\theta(x_t, t) \right), \beta_t \mathbf{I} \right),
\end{equation}
where $\alpha_t = 1 - \beta_t$, $\bar{\alpha}_t = \prod_{i=1}^{T} \alpha_i$, and $T$ represents the total number of timesteps. Notably, $\epsilon_\theta$ is a denoising network parameterized by $\theta$, which takes $x_t$ and $t$ as inputs to predict the noise needed for denoising. In an image generation procedure using $\mathcal{T}$ timesteps, $\epsilon_\theta$ must perform inference $\mathcal{T}$ times, which constitutes the majority of computational costs in diffusion models. Recent studies have shown that implementing $\epsilon_\theta$ as a transformer often improves the quality of generation.

\begin{figure*}
    \centering
    
    \includegraphics[width=\linewidth]{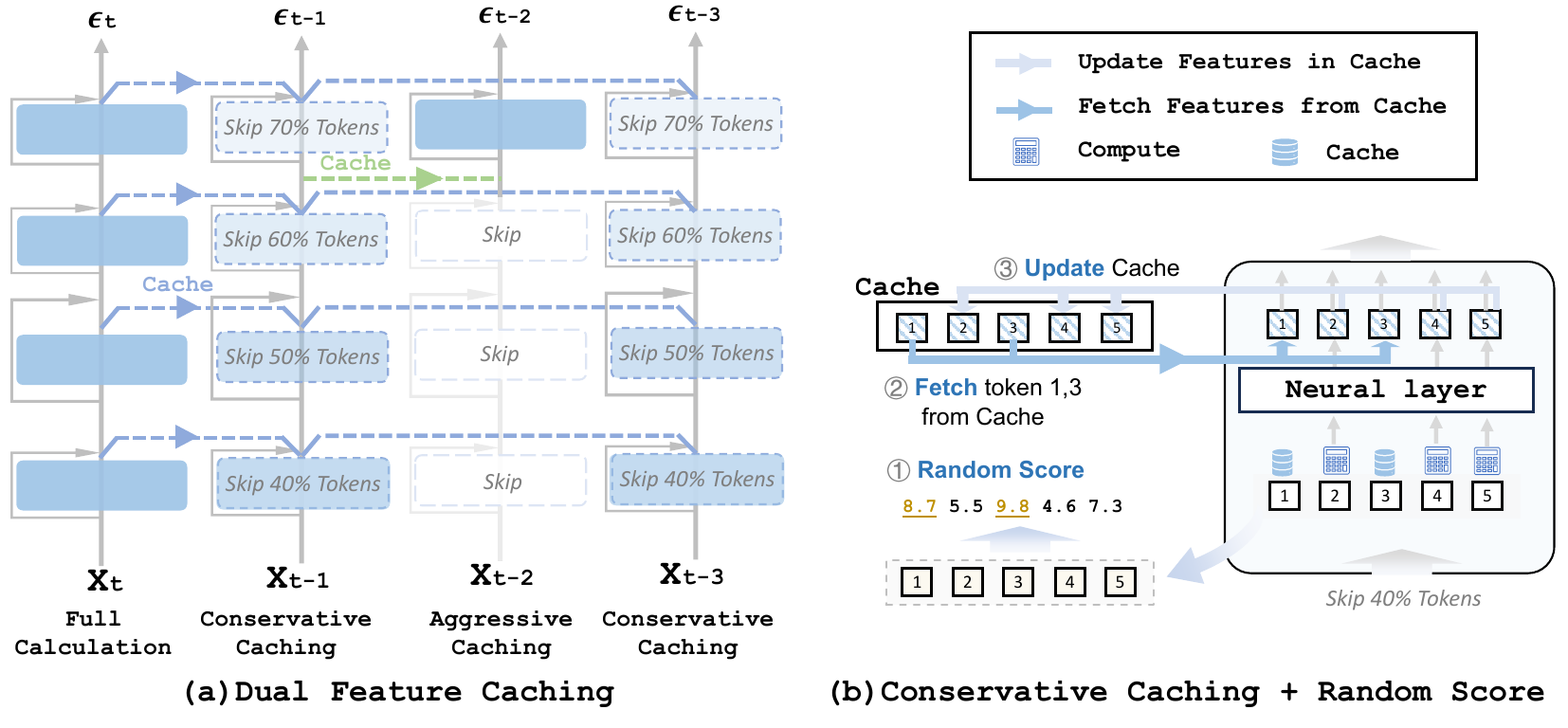}
    \caption{The overview of \texttt{\texttt{DuCa}}. (a) In the first timestep of a cache cycle, \texttt{\texttt{DuCa}} computes all tokens and stores them in the cache as initialization. Then, \texttt{\texttt{DuCa}} performs a conservative caching step followed by an aggressive caching step, and then another conservative caching step, alternating in this manner to form a complete cache cycle. (b) The proposed method selects tokens for \texttt{\texttt{DuCa}} by setting random scores for all tokens to diversity. The tokens with the largest random scores are selected as cached tokens. The computation results for these tokens are directly retrieved from the cache, while the remaining tokens are computed through the neural layer, with their updated results stored back into the cache. }
    \label{fig:Method}
    \vspace{-3mm}
\end{figure*}

\paragraph{Diffusion Transformers}
Diffusion transformers~(DiT) are usually composed of stacking groups of self-attention layers $\mathcal{F}_{\text{SA}}$, multilayer perceptron $\mathcal{F}_{\text{MLP}}$, and cross-attention layers $\mathcal{F}_{\text{CA}}$ (for conditional generation). It can be roughly formulated as:\vspace{-2mm}
\begin{equation}
\label{eq: DiT}
\vspace{-1mm}
\begin{aligned}
    &\mathcal{G} = g_1 \circ g_2 \circ \dots \circ  g_L, ~~~\text{where}\\
    &g_l = \mathcal{F}_{\text{SA}}^{l} \circ \mathcal{F}_{\text{CA}}^{l} \circ \mathcal{F}_{\text{MLP}}^{l},  \\
\end{aligned}
\end{equation}
$\mathcal{G}$, $g_l$, and $\mathcal{F}^{l}$ represent the DiT model, the blocks in DiT, and different layers in a DiT block, respectively. $l$ denotes the index of DiT blocks, and $L$ denotes the depth of DiT.
In diffusion transformers, the input data $\mathbf{x}_t$ consists of a sequence of tokens representing different patches in the generated images, expressed as $\mathbf{x}_t=\{x_i\}_{i=1}^{H\times W}$, where $H$ and $W$ indicate the height and width of the images or their latent code, respectively. 
Then, the computation between $\mathcal{F}$ and $\mathbf{x}$ can be generally written as $\mathcal{F}(\mathbf{x}) = \mathbf{x} + \text{AdaLN}\circ f(\mathbf{x})$, where the first item $\mathbf{x}$ indicates the residual connection and $f$ denotes the MLP or attention layers. AdaLN indicates the adaptive layer normalization~\cite{Guo2022AdaLNAV}.  

\noindent\textbf{Feature Caching:} Feature caching aims to reuse the features that are computed and cached in the past timesteps to skip the computation in future timesteps. 
It is usually performed in multiple repeated caching cycles. Assuming a caching cycle with $N$ timesteps from the timestep $t$ to $t+N-1$, then feature caching performs the full computation in the timestep $t$ and stores the obtained features in a cache.
This progress can be denoted as $\mathcal{C}[l]:=g_l(x_t)$, where $\mathcal{C}$ denotes the cache and ``:='' denotes assigning value operation. Then, in timesteps from $t+1$ to $t+N-1$, the computation in these steps is skipped by reusing the cached features, which can be formulated as $g_l(x_{t+i}):=\mathcal{C}[l], i\leq N-1$.
For simplicity, in this paper, the timestep $t$ and the timesteps in $[t+1, t+N-1]$ are named the freshing step and the \emph{caching} steps, respectively.

\subsection{Dual Feature Caching}

The proposed dual feature caching aims to perform aggressive caching and conservative caching alternatively in the caching steps. Hence, in this section, we first introduce the details of the two caching strategies.
\paragraph{Aggressive Caching in \texttt{\texttt{DuCa}}} As introduced in Figure~\ref{fig:aggressive-conservative}(b), aggressive caching directly replaces the output of the DiT block with the cached feature which is computed in previous layers, which means that all computations of the first $l$ layers can be skipped.
This can be expressed as $g_l:= \mathcal{C}_{}[l]$, where $\mathcal{C}_{}[l]$ denotes the cached feature at the $l_{th}$ layer.
Then, the computation with aggressive caching can be written as: \vspace {-3mm}
\begin{equation}
\label{eq: Aggressive-Caching}
    \mathcal{G}(x)_{t}:=  g_{l+1} \circ ...\circ g_L^{t}(\mathcal{C}[l]).
    \vspace{-3mm}
\end{equation}

To achieve the maximal acceleration with aggressive caching, we set $l = L-1$ in most of our experiments, which indicates almost skipping all the layers in the caching step.
\paragraph{Conservative Caching in \texttt{\texttt{DuCa}}} As shown in Figure \ref{fig:aggressive-conservative}(a), conservative caching does not cache and reuse all the features. Instead, it caches and reuses only the features without the residual connection, which allows it to allocate the computation to some layers in the caching steps to correct the cached features.
Recently, \texttt{ToCa} \cite{zou2024accelerating} introduced a token-wise approach that selectively reuses features based on the importance of each token and its sensitivity to caching, which enables us to perform computation only in the important tokens. Following \texttt{ToCa}, we first calculate an importance score $\mathcal{S}$ for each token $x_i$ and then use this score to divide the index set of tokens $\mathcal{I}$ into two \textit{complementary} sets: $\mathcal{I}_{\text{Cache}}$, for tokens to be cached, and $\mathcal{I}_{\text{Compute}}$, for tokens to be computed. Subsequently, the computation for the $i_{th}$ token $x_i$ on the $l_{th}$ layer can be expressed as: 
  $\gamma_i f^{l}(x_i) + (1-\gamma_i) \mathcal{C}[l](x_i),$
where $\gamma_i=0$ for $i \in \mathcal{I}_{\text{Cache}}$ and $\gamma_i=1$ for $i \in \mathcal{I}_{\text{Compute}}$. Here, $\mathcal{C}[l](x_i)$ denotes retrieving the cached value of $x_i$ from the conservative cache $\mathcal{C}$ at layer $l$, which incurs no computation cost, enabling partial omission of calculations. After this selective computation in the token dimension, the computed features are updated to $\mathcal{C}[l]$ for future reusing.

\paragraph{Key Issue of Attention-Based Selection Methods}
As the previous token-wise conservative caching method, \texttt{ToCa} offers two attention map-based selection strategies that have shown good effectiveness. However, the attention map-based selection strategy is incompatible with another commonly used attention-optimization method  FlashAttention\cite{Dao2022FlashAttentionFA} and memory-efficient attention, which can not provide the attention scores but are capable of reducing the memory costs of attention from $O(N^2)$ to $O(N)$ and significant accelerating the attention layers. This conflict between attention scores-based token selection with this optimized attention computation makes \texttt{ToCa} less practical in application and raises the requirements for attention score-free token selection methods.

  \begin{table}[t]
    \caption{Token selection methods for FLUX on Image Reward. The experiments were conducted on the FLUX.1-dev model with a full activation interval of 4, selecting $10\%$ of tokens at each cache step. A higher Image Reward means better performance.
    }
    \vspace{-3mm}
    \centering
    \small
    \resizebox{0.48\textwidth}{!}{
      \begin{tabular}{l | c | c | c}
        \toprule
        {\bf Score}     &{\bf Efficient Attention} &{\bf Max$^\spadesuit$} &{\bf Min$^\heartsuit$} \\
        \hline
        {Attention} & \ding{56}                & 0.9798                & 0.9707 \\
        {K-norm}    & \ding{52}                & 0.9783                & 0.9726 \\
        {V-norm}    & \ding{52}                & 0.9747                & 0.9798 \\
        {Similarity}& \ding{52}                & 0.9715                & 0.9811 \\
        \hline
        {Random}    & \ding{52}                &\multicolumn{2}{c}{0.9806}      \\
        {Original Model}  & \ding{52}                &\multicolumn{2}{c}{0.9898}      \\
        \bottomrule
      \end{tabular}
      }
      
      \label{table:token-selection}
      \vspace{-2mm}
  \end{table}
\paragraph{Diversity matters more in token selection} As shown in Table \ref{table:token-selection}, we compare attention scores with other classic token selection methods used in large language and multimodal models, such as K-norm and V-norm, which select tokens based on maximizing and minimizing the corresponding values, respectively. We then compare these results with the simplest and most naive random selection. Surprisingly, while the method of selecting tokens based on the highest attention score in \texttt{ToCa} is the best-performing approach among all standard-based selection methods, it still performs slightly worse than random selection. 

The results suggest that rather than the traditional assumption that tokens can be distinguished as important or unimportant, selecting a diverse and broad range of tokens is more beneficial for performance. Based on this hypothesis, we designed an experiment that scores tokens based on similarity. For each cache step in Table \ref{table:token-selection}, $10\%$ of all tokens were selected. First, $1\%$ of all tokens were randomly chosen as the base tokens. Then, for the remaining $9\%$, tokens were selected based on two criteria: one method selects tokens with the highest similarity to the base tokens, aiming to choose tokens of the same type, while the other method selects those with the lowest similarity, encouraging greater diversity in the selected tokens. The method based on maximum similarity achieved the worst performance among all selection methods, while the method based on minimum similarity achieved the best performance, further confirming the \textit{importance of diversity in token selection}. Considering that the difference between the minimum similarity-based method and the random selection method is small and it does not require additional computation, the random token selection method is recommended.

\subsection{Overall Framework}

Based on the understanding of aggressive and conservative caching,  \texttt{\texttt{DuCa}} introduces a caching scheme that alternates conservative and aggressive caching in different steps. Random selection serves as the selection strategy for conservative caching, as shown in Figure~\ref{fig:Method}.
This design leverages the realignment capability of token-wise conservative caching to correct the misaligned timestep information occurring in aggressive caching steps. Therefore, it enables the high local compression rate of aggressive caching to be applied to more steps while further reducing redundant computations in conservative caching steps. We arrange token-wise conservative caching steps on odd-numbered steps following a fresh step and aggressive caching steps on even-numbered steps. 




\begin{table*}[t]
    \caption{\textbf{Quantitative comparison in text-to-image generation} for FLUX on Image Reward.
    }\vspace{-3mm}
    \centering
    \setlength\tabcolsep{7.0pt} 
      \small
      \resizebox{\textwidth}{!}{\begin{tabular}{l | c | c  c | c  c | c | c}
        \toprule
        {\bf Method} & {\bf Efficient} &\multicolumn{4}{c|}{\bf Acceleration} &{\bf Image Reward $\uparrow$} &\bf Geneval$\uparrow$ \\
        \cline{3-6}
        {\bf FLUX.1\cite{flux2024}} & {\bf Attention \cite{Dao2022FlashAttentionFA}} & {\bf Latency(s) $\downarrow$}  & {\bf Speed $\uparrow$} & {\bf FLOPs(T) $\downarrow$} & {\bf Speed $\uparrow$} & \bf DrawBench &\bf Overall \\
        \midrule
      
      $\textbf{[dev]: 50 steps}$ 
                               & \ding{52}  &  {17.20}  & {1.00$\times$} & {3719.50}   & {1.00$\times$} & {0.9898}  &{0.6752}      \\ 
      \midrule

      {$60\%$\textbf{ steps}}  & \ding{52}  &  {10.49}   & {1.64$\times$} & {2231.70}  & {1.67$\times$} & {0.9739}  &{0.6700}      \\
      {$50\%$\textbf{ steps}}  & \ding{52}  &  {8.80}  & {1.95$\times$} & {1859.75}   & {2.00$\times$} & {0.9429}  &{0.6655}             \\
      {$40\%$\textbf{ steps}}  & \ding{52}  &  {7.11}   & {2.42$\times$} & {1487.80}  & {2.62$\times$} & {0.9317}  &{0.6606}             \\

     {$\Delta$-DiT} ($\mathcal{N}=2$) & \ding{52}  &  {11.86}  & {1.45$\times$} & {2480.01}   & {1.50$\times$} & {0.9444}    &{0.6658}   \\
    {$\Delta$-DiT} ($\mathcal{N}=3$) & \ding{52}  &  {8.68}  & {1.98$\times$} & {1686.76}   & {2.21$\times$} & {0.8721}    &{0.6413}   \\

     $\textbf{TeaCache}$ 
      
                               & \ding{52}  &  {7.41}     & {2.32$\times$} & {1114.44}  & {2.63$\times$} & {0.9344}   &{0.6722}  \\
      $\textbf{FORA}$ 
      \cite{selvaraju2024fora}& \ding{52}  &  {7.08}  & {2.43$\times$} & {1320.07}   & {2.82$\times$} & {0.9227}    &{0.6594}   \\
      $\textbf{ToCa}$ 
      \cite{zou2024accelerating} 
                               & \ding{56}  &  {10.80}     & {1.59$\times$} & {1126.76}  & {3.30$\times$} & {0.9731}   &{0.6750}  \\

      \rowcolor{gray!20}
      $\textbf{\texttt{DuCa}} $ $(\mathcal{N}=4,R=90\%)$ 
                               & \ding{52}  &  {6.22}   & {2.77$\times$} & {1190.24}   & {3.13$\times$} &\underline{0.9895}   &\underline{0.6750}  \\
      \rowcolor{gray!20}
      $\textbf{\texttt{DuCa}} $ $(\mathcal{N}=5,R=90\%)$ 
                               & \ding{52}  &  \underline{5.88}   & \underline{2.93$\times$}  & \underline{1078.34}  & \underline{3.45$\times$} &\textbf{0.9896}   &{0.6747}  \\
      \rowcolor{gray!20}
      $\textbf{\texttt{DuCa}} $ $(\mathcal{N}=6,R=90\%)$ 
                               & \ding{52}  &  \textbf{5.14}    & \textbf{3.35$\times$} & \textbf{917.32}  & \textbf{4.05$\times$} &{0.9654}   &\bf{0.6756}  \\
      %
      %
        \bottomrule
      \end{tabular}
      }
      
      \label{table:FLUX-Metrics}
  \end{table*}

\section{Experiments}

\subsection{Experiment Settings}

\subsubsection{Model Configurations}

\label{sec:model-configurations}
Experiments are conducted on four widely known DiT-based models on the three generation tasks with the correlating default samplers, including the FLUX.1-dev model \cite{flux2024} with 50 rectified flow sampling steps, HiDream model with 50 rectified flow sampling steps, OpenSora \cite{opensora} with 30 rectified flow sampling steps \cite{refitiedflow} for text-to-video generation, and DiT-XL/2 \cite{peebles2023dit} with 50 DDIM \cite{songDDIM} steps for class-conditional image generation. All the experiments are conducted with NVIDIA H800 80GB GPUs. Additionally, we have supplemented our experiments with the smaller text-to-image generation model PixArt-$\alpha$ \cite{chen2023pixartalpha} with 20 DPM Solver++ \cite{lu2022dpm++} steps and included the detailed design of all experiments. More details are available in the appendix.


\subsubsection{Evaluation and Metrics}
\noindent\textbf{Text-to-image generation} For the FLUX.1-dev and HiDream model with a high resolution of $1000 \times 1000$, we utilized 200 high-quality prompts from DrawBench~\cite{Saharia2022PhotorealisticTD}, which focus on generating various types of images. We then generated images using these prompts and employed Image Reward~\cite{xuImageRewardLearningEvaluating2023} to assess the quality of the generated images while using the GenEval~\cite{Ghosh2023GenEvalAO} benchmarks to evaluate the alignment between the images and the text.

\noindent\textbf{Text-to-video generation} \textbf{VBench} \cite{huang2024vbench} evaluation framework is applied to evaluate the performance of the text-to-video model. With 5 videos per prompt and 950 prompts from the framework, 4,750  480p, 9:16, and 2s videos are generated to evaluate the 16 aspects proposed by VBench. 

\begin{table}[htbp]
\vspace{-11pt}
\caption{ Quantitative results for HiDream-l1.}\label{tab:HiDream}
\vspace{-10pt}
\resizebox{.49\textwidth}{!}{
\begin{tabular}{l|cc|cc|c}
\toprule
        {\bf Method}  &\multicolumn{4}{c|}{\bf Acceleration} &{\bf Image Reward $\uparrow$} \\ 
        \cline{2-5}
        HiDream-l1     & {\bf Latency(s) $\downarrow$}  & {\bf Speed $\uparrow$} & {\bf FLOPs(T) $\downarrow$} & {\bf Speed $\uparrow$} & \bf DrawBench \\ \midrule
$\textbf{[full]: 50 steps}$      & 48.81 & $\times 1.00$ & 7780.0 & $\times 1.00$  & 1.182  \\
\midrule
25 steps     & 24.41 & $\times 2.00$ & 3890.0 & $\times 2.00$  & 1.128  \\
ToCa ($N=4$) & 31.22 & $\times 1.56$ & 2540.1 & $\times 3.06$  & 1.111  \\
PAB  ($N=8$) & 36.33 & $\times 1.34$ & 5962.0 & $\times 1.30$  & 0.8876  \\
\rowcolor{gray!20}
\texttt{\textbf{DuCa}} ($N=3, R=90\%$) & 19.88 & $\times 2.46$ & 2873.4 & $\times 2.71$  & 1.152  \\
\rowcolor{gray!20}
\texttt{\textbf{DuCa}} ($N=4, R=90\%$) & 15.36 & $\times 3.18$ & 2197.3 & $\times 3.54$  & 1.090  \\

\bottomrule
\end{tabular}}
\vspace{-11pt}
\end{table}

\noindent\textbf{Class-conditional image generation} 50,000 $256 \times 256$ images uniformly sampled from the 1,000 classes in ImageNet\cite{deng2009imagenet} are generated to evaluate the FID-50k \cite{heusel2017fid}. Besides, sFID, Precision, and Recall are introduced as supplementary metrics.

\subsubsection{Results on text-to-image generation}

The generation results of the proposed \texttt{\texttt{DuCa}} under various configurations are compared against recent state-of-the-art acceleration methods, including \texttt{ToCa} \cite{zou2024accelerating}, FORA \cite{selvaraju2024fora}, and a series of methods that directly reduce sampling steps. These comparisons are conducted on FLUX.1-dev~\cite{flux2024} and HiDream at a resolution of $1000 \times 1000$, as detailed in Table \ref{table:FLUX-Metrics} and \ref{tab:HiDream}.

\textbf{In terms of \textit{generation quality}}, the quantitative results on the Image Reward indicate that \texttt{ToCa} achieves a maximum computational compression (\emph{i.e.}, FLOPs speedup) of 3.30$\times$, with an Image Reward of 0.9731, representing a decrease of 0.0167 compared to the original model. In contrast, \texttt{\texttt{DuCa}} achieves an Image Reward score of 0.9896 under a computational compression of 3.45$\times$, remaining almost unchanged compared to the original model, reducing the quality loss to $1.2\%$ of that observed with \texttt{ToCa}.

\textbf{From the perspective of text-image alignment}, the quantitative results on the GenEval metric demonstrate that \texttt{\texttt{DuCa}} maintains nearly lossless performance compared to the original model. Even under a computational compression of up to 4.05$\times$, \texttt{\texttt{DuCa}} still ensures excellent alignment capability. In contrast, FORA exhibits a noticeable decline in alignment ability at a compression ratio of $2.82\times$.

\begin{table*}[t]
\vspace{-0.3cm}
    \caption{\textbf{Quantitative comparison on class-to-image generation} on ImageNet with \text{DiT-XL/2.} }
    \vspace{-0.2cm}
    \centering
    \setlength\tabcolsep{7.7pt} 
      \small
       \resizebox{0.92\textwidth}{!}{\begin{tabular}{l | c c c | c c c c }
        \toprule
        \bf Method  & \bf Latency(s) $\downarrow$ & \bf FLOPs(T) $\downarrow$ & \bf Speed $\uparrow$  & \bf sFID $\downarrow$ & \bf FID $\downarrow$ & \bf Precision $\uparrow$& \bf Recall $\uparrow$  \\
        \midrule
      {\textbf{$\text{DDIM-50 steps}$}} & {0.417}  & {23.74}  & {1.00$\times$}  &  {{4.29}} &  {{2.31}}  & {0.80} &  {0.59} \\
      {\textbf{$\text{DDIM-25 steps}$}} & {0.209}  & {11.87}  & {2.00$\times$}  &  {{4.63}} &  {{3.07}}  & {0.79} &  {0.58} \\
      {\textbf{$\text{DDIM-20 steps}$}} & {0.167}  & {9.49}  & {2.50$\times$}  &  {{5.04}} &  {{3.69}}  & {0.78} &  {0.58} \\
      {\textbf{$\text{DDIM-17 steps}$}} & {0.142}  & {8.07}  & {2.94$\times$}  &  {{5.65}} &  {{4.48}}  & {0.77} &  {0.56} \\
      \midrule     
       {$\textbf{FORA} $} \cite{selvaraju2024fora}     & 0.197 & 8.57 & {2.77$\times$}  & 6.32 & 3.76 & 0.79 & 0.54 \\
      $\Delta$-DiT ($\mathcal{N}=2$)     & 0.246 & 18.04 & {1.31$\times$}  & 4.67 & 2.69 & 0.79 & 0.56 \\
      $\Delta$-DiT ($\mathcal{N}=3$)     & 0.173 & 16.14 & {1.92$\times$}  & 5.7 & 3.75 & 0.79 & 0.54 \\
      
      {\textbf{\texttt{ToCa}(a)}} & {0.195} & {8.73} & {2.72$\times$} & {4.98}  & {3.31} & {0.79} & {0.55} \\
      {\textbf{\texttt{ToCa}(b)}} & {0.214} & {10.23} & {2.32$\times$} & {4.73}  & {2.88} & {0.79} & {0.57} \\
      \rowcolor{gray!20}

      \rowcolor{gray!20}
      {\textbf{\texttt{DuCa}(a)}} & \textbf{0.177} & \textbf{8.76} & {2.71$\times$} & {4.53}  & {3.00} & {0.80} & {0.57} \\
      \rowcolor{gray!20}
      {\textbf{\texttt{DuCa}(b)}} & {0.198} & {9.58} & {2.48$\times$} & \bf{4.59}  & \bf{2.84} & \bf{0.80} & \bf{0.58} \\
        \bottomrule
      \end{tabular}}
      
      \label{table:main_ldm_imagenet}
      \vspace{-3mm}
  \end{table*}

\subsection{Quantitative and Qualitative Results}
\begin{figure*}
    \centering
    \includegraphics[width=0.95\linewidth]{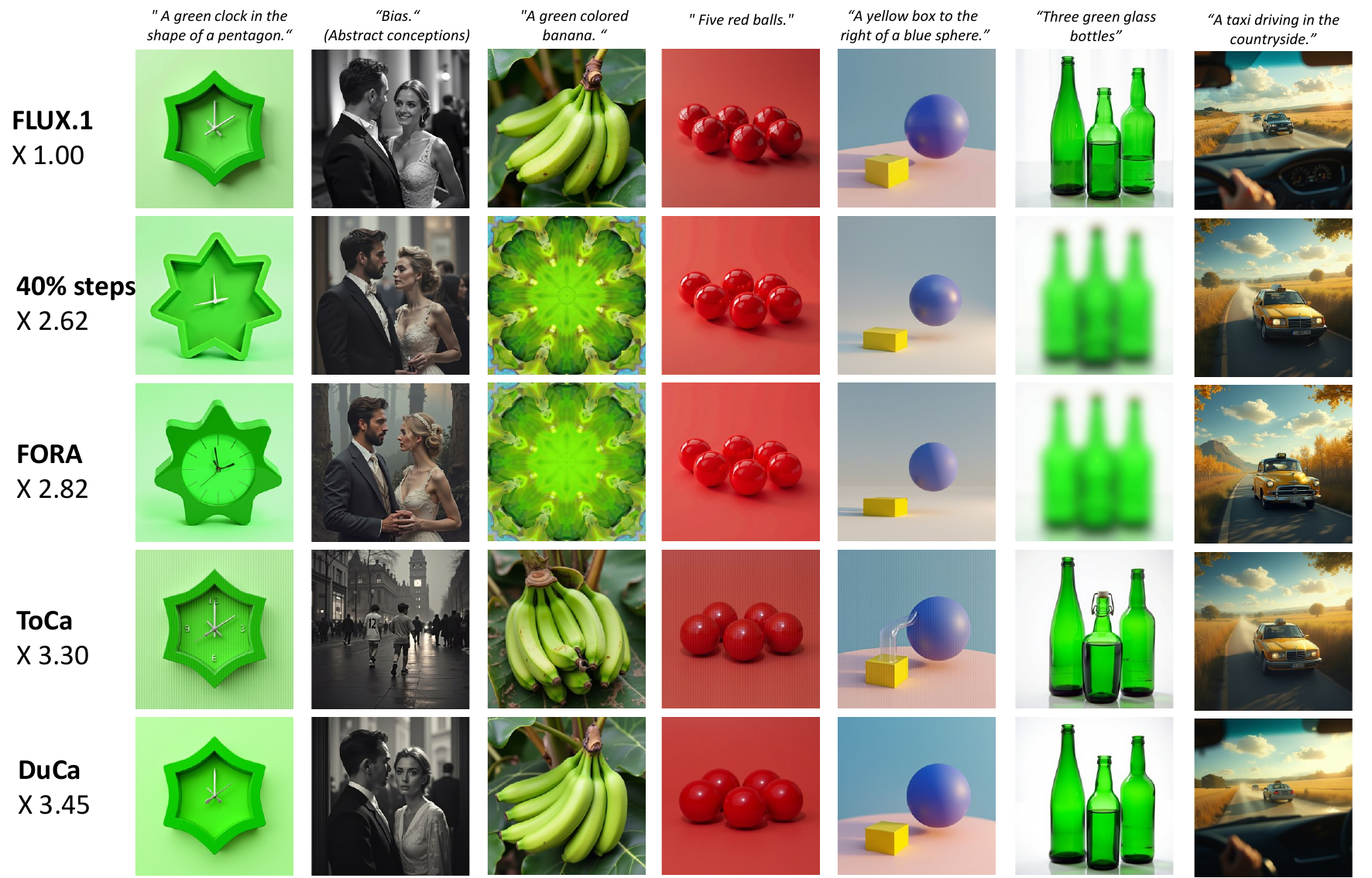}
    \vspace{-3mm}
    \caption{Visualization results for different acceleration methods on FLUX.1-dev model.}
    \vspace{-3mm}
    \label{fig:Visualization-FLUX}

\end{figure*}

\textbf{From a practical application perspective}, although the computational compression ratio of the proposed \texttt{\texttt{DuCa}} at 3.45× is not significantly different from \texttt{ToCa}'s 3.30$\times$, \texttt{\texttt{DuCa}} not only further enhances generation quality but also achieves a latency speedup ratio of 2.93 due to its compatibility with FlashAttention~\cite{Dao2022FlashAttentionFA}. This far surpasses \texttt{\texttt{ToCa}}'s 1.59$\times$ speedup, making \texttt{\texttt{DuCa}} superior to the current state-of-the-art \texttt{\texttt{ToCa}} in both speed and generation quality.

In Figure \ref{fig:Visualization-FLUX}, we further present the generation results of various acceleration methods. From Figure \ref{fig:Visualization-FLUX}, it can be observed that across multiple samples, \texttt{DuCa}'s generated images are closer to those of the original model without acceleration compared to other methods. The 40$\%$ steps and FORA schemes not only offer lower acceleration ratios but are also more prone to visual artifacts. Although \texttt{\texttt{ToCa}} exhibits better details to some extent, we notice that in the 1st, 4th, and 5th columns from left to right, there are inexplicable grid-like distortions. In contrast, \texttt{\texttt{DuCa}} emerges as a superior choice in terms of both speed and performance.

\subsubsection{Results on text-to-video generation}

The generation results of the state-of-the-art acceleration methods $\Delta$-DiT, T-GATE\cite{tgatev1}, PAB\cite{zhao2024PAB}, 15 rflow sampling steps, \texttt{\texttt{ToCa}}\cite{zou2024accelerating}, FORA \cite{selvaraju2024fora} and proposed \texttt{\texttt{DuCa}} on the text-to-video model OpenSora\cite{opensora} are shown in Table \ref{table:main_ldm_OpenSora}. 

As shown in Table \ref{table:main_ldm_OpenSora}, \texttt{\texttt{DuCa}} achieves a further acceleration from 2.36 $\times$ (as with \texttt{\texttt{ToCa}}) to 2.50 $\times$ with virtually no loss in generation quality. Additionally, despite having the lowest computational cost among all the compared methods, \texttt{\texttt{DuCa}} still achieves the highest VBench score of 78.83, demonstrating that the proposed strategy in \texttt{\texttt{DuCa}}, which alternates between aggressive and conservative caching steps, effectively eliminates the redundant computations in the fully conservative caching approach.

\begin{figure*}
    \centering
    \includegraphics[width=0.98\linewidth]{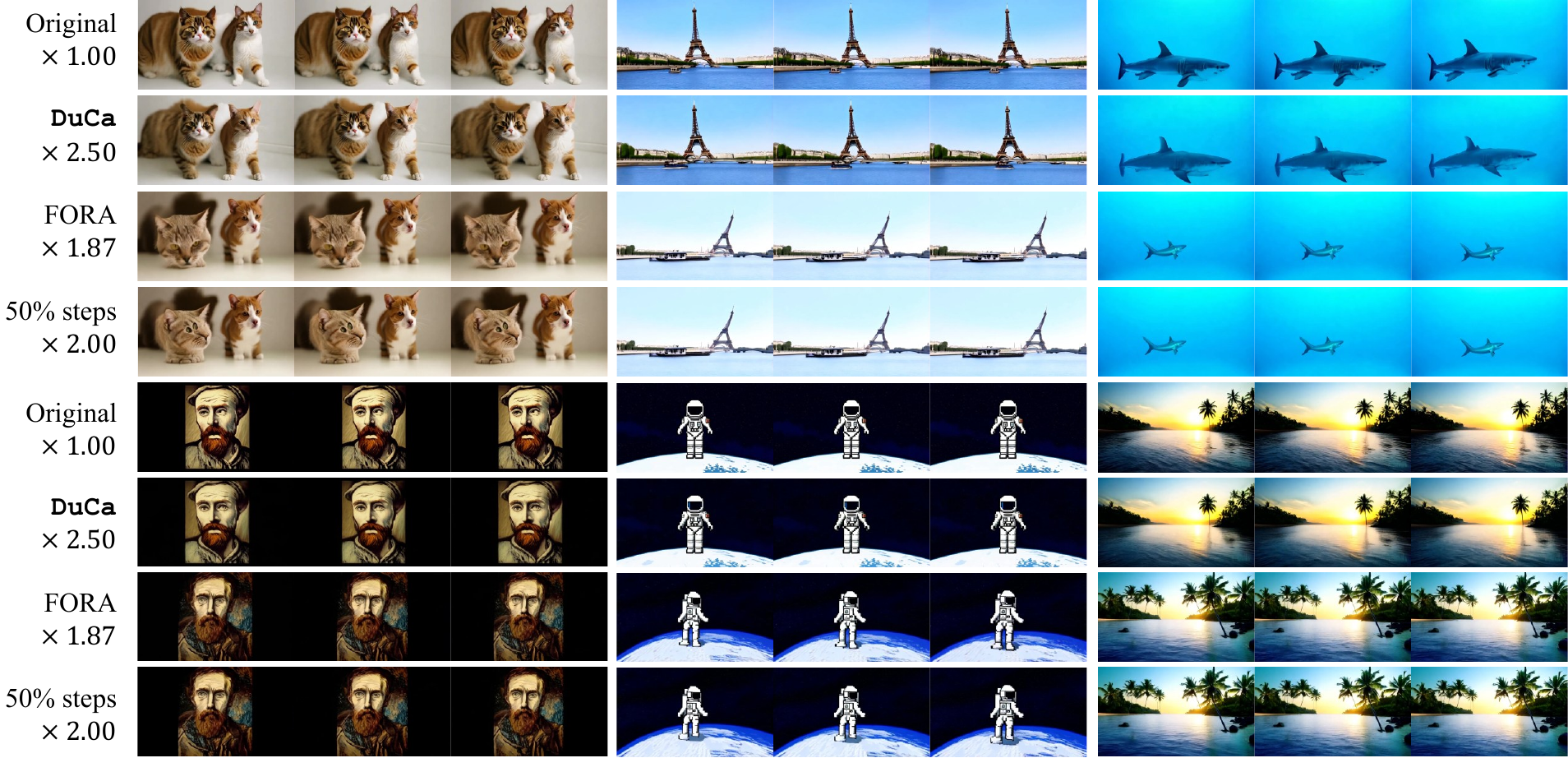}
     \vspace{-1mm}
    \caption{Visualization results for different methods on OpenSora (videos available in the supplementary material).}
    \label{fig:Visualization-OpenSora}
    \vspace{-2mm}
\end{figure*}

\begin{table}[t]
    \caption{\textbf{Quantitative comparison in text-to-video generation} on VBench. *Results  are from PAB \cite{zhao2024PAB}. PAB$^{1-3}$ indicate PAB with different hyper-parameters.
    }
    \vspace{-1pt}
    \centering
    \setlength\tabcolsep{1.2pt} 
      \small
      \resizebox{0.48\textwidth}{!}{
      \begin{tabular}{l | c | c | c | c }
        \toprule
        {\bf Method} & {\bf Latency(s)$\downarrow$} & {\bf FLOPs(T)$\downarrow$} & {\bf Speed$\uparrow$} & \bf VBench(\%)$\uparrow$ \\
        \midrule
      $\textbf{OpenSora}$  \cite{opensora}   &  54.83  & 3283.20   & {1.00$\times$} & 79.41       \\ 
      \midrule
      {$\textbf{$\Delta$-DiT}^*$}  \cite{chen2024delta-dit} & 53.45  & 3166.47  & {1.04$\times$} & 78.21
       \\
      {$\textbf{T-GATE}^*$} \cite{tgatev1}  & 45.91     & 2818.40  & {1.16$\times$} & 77.61
       \\
      {\textbf{PAB$_{}^{1*}$}}  \cite{zhao2024PAB} &  41.05   & 2657.70  & {1.24$\times$} & 78.51          
       \\
      {\textbf{PAB$_{}^{2*}$}} \cite{zhao2024PAB}  &  39.96   & 2615.15  & {1.26$\times$} & 77.64          
       \\
      {\textbf{PAB$_{}^{3*}$}}  \cite{zhao2024PAB} &  38.26   & 2558.25  & {1.28$\times$} & 76.95    
       \\
      \midrule
      {$50\%$\textbf{ steps}}   &  28.85   & 1641.60   & {2.00$\times$} & 76.78             \\
      {$\textbf{FORA} $} \cite{selvaraju2024fora} & 33.19 & 1751.32   & {1.87$\times$} & 76.91             \\
      {\textbf{\texttt{ToCa}}}\cite{zou2024accelerating} & 29.03 & 1394.03  & {2.36$\times$} & {78.59}    \\
      \rowcolor{gray!20}
      {\textbf{\texttt{DuCa}}} &\bf{19.16} & \bf{1315.62}  &  \bf{2.50$\times$} & \textbf{78.83} \\
        \bottomrule
      \end{tabular}}
      
      \label{table:main_ldm_OpenSora}
      \vspace{-4pt}
      \vspace{-3mm}
  \end{table}

The advantages of \texttt{DuCa}, stemming from its diverse selection strategies, are also evident in its comparison with \texttt{\texttt{ToCa}}: although \texttt{\texttt{DuCa}} eliminates unnecessary redundant computations, resulting in lower actual computational load than \texttt{\texttt{ToCa}}, its VBench score decreases by only 0.58 compared to the original model, while \texttt{\texttt{ToCa}}'s score drops by 0.82, indicating that \texttt{\texttt{DuCa}} reduces quality loss by $29.3\%$.

Furthermore, thanks to \texttt{DuCa}'s compatibility with FlashAttention~\cite{Dao2022FlashAttentionFA}, the generation of a single 480p, 2s video takes only 19.16s with \texttt{DuCa}, compared to 29.03s with \texttt{\texttt{ToCa}}, resulting in a speedup of nearly 10s.

\begin{figure}
    \centering
    \includegraphics[width=0.85\linewidth]{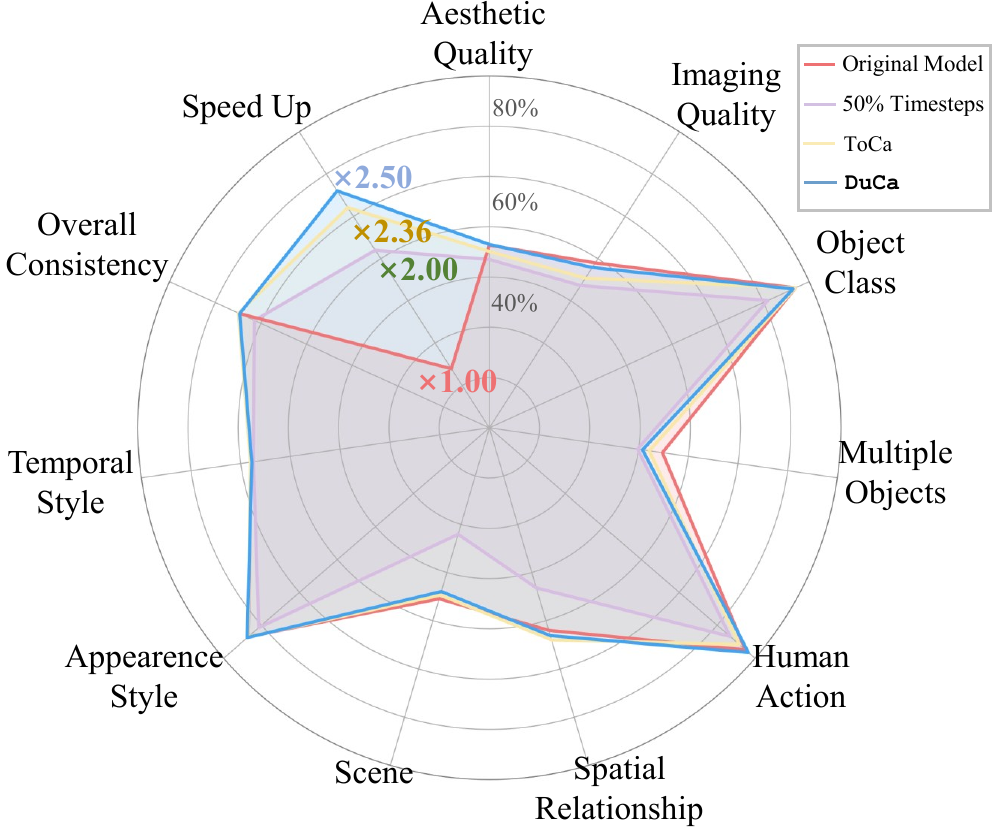}
    \vspace{-3mm}
    \caption{VBench metrics and acceleration ratio of proposed \textbf{\texttt{\texttt{DuCa}}} and other methods.}
    \label{fig:opensora-radar}
    \vspace{-5mm}
\end{figure}

In Figure \ref{fig:opensora-radar}, we further present the performance of \texttt{\texttt{DuCa}} and the aforementioned methods across various dimensions of the VBench metric. \texttt{\texttt{DuCa}} shows results consistent with the non-accelerated original OpenSora model across multiple metrics. As shown in  Figure \ref{fig:Visualization-OpenSora}, it is evident that \texttt{DuCa} is closer to the original model compared to other methods.

\subsubsection{Results on class-conditional image generation}
In Table \ref{table:main_ldm_imagenet}, we compare \texttt{\texttt{ToCa}}\cite{zou2024accelerating}, FORA \cite{selvaraju2024fora}, the direct reduction of the corresponding DDIM steps and the proposed \texttt{\texttt{DuCa}}. The experimental results indicate that with comparable FID scores, representing generation quality, \texttt{\texttt{DuCa}} further enhances acceleration. For instance, \texttt{\texttt{DuCa}}(b) with Self-Attention score achieves a similar FID-50k as \texttt{\texttt{ToCa}}(b), but the acceleration ratio increases from 2.32 to 2.48. Additionally, under a higher acceleration scenario with an acceleration ratio around 2.7, \texttt{\texttt{DuCa}} achieves an FID-50k of 3.00, further reducing the FID-50k by 0.31 compared to \texttt{\texttt{ToCa}}'s 3.31, demonstrating that \texttt{\texttt{DuCa}} effectively reallocates computation.

\begin{figure*}
    \centering
    \includegraphics[width=\linewidth]{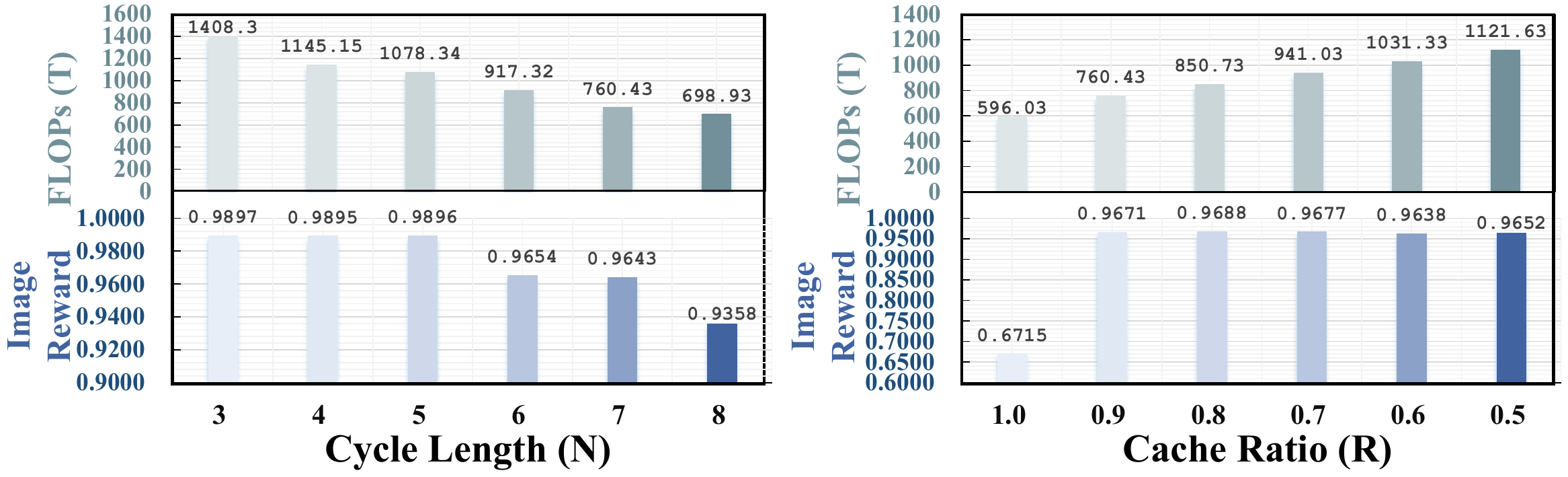}

    \caption{Ablation study on cycle length $N$ and cache ratio $R$. Setting $N=5$ and $R=0.9$ yields the best trade-off between computation and quality.}
    \label{fig:Ratio-ablation}

\end{figure*}

\subsection{Ablation Studies}
\begin{table}[t]
    \caption{{Different caching methods on FLUX.1-dev.} 
    }
    \vspace{-2mm}
    \centering
      \small
        \resizebox{0.48\textwidth}{!}{\begin{tabular}{l | c | c }
        \toprule
        {\bf Methods} & \bf {Image Reward} & {\bf FLOPs(T) $\downarrow$}  \\

        \midrule

      \textbf{Conservative-only}      & {0.9789} & {1263.22}  \\
      \textbf{Aggressive-only}       & {0.8539} & {1042.12}   \\
      \textbf{\texttt{DuCa}}         & \textbf{0.9895} & {1190.24} \\

        \bottomrule
      \end{tabular}}

      \label{table: Ablation-Aggressive-Conservative}
  \end{table}

\paragraph{Diversity in Token Selection}
As demonstrated in Table \ref{table:token-selection}, the results prove that the adopted random selection method, due to its consideration of diversity in token selection, outperforms a series of traditional methods based on specific metrics while maintaining compatibility with the Efficient Attention framework. Additionally, through experiments designed around similarity, we have explored and strongly validated the importance of diversity in token selection.

As shown in Table \ref{table: Ablation-Aggressive-Conservative}, we compare the conservative-only, aggressive-only, and alternating \texttt{\texttt{DuCa}} structures, where the aggressive structure is used in activated odd steps and the conservative structure in even steps, on FLUX.1-dev in terms of Image Reward and FLOPs. The conservative-only structure appears to have good generation quality but incurs an excessively high computational cost. In contrast, aggressive-only reduces computation to nearly half of conservative-only but suffers significantly in generation quality. Our proposed \texttt{\texttt{DuCa}} structure achieves a balanced computational load between the two, reducing the computational cost compared to the conservative-only method while maintaining comparable or even better generation quality.

\paragraph{Choice of Cycle Length and Cache Ratio}
We further conduct ablation experiments to investigate the influence of cycle length ($N$) and cache ratio ($R$) on performance, as illustrated in Figure~\ref{fig:Ratio-ablation}. When varying $N$ from 3 to 8, we observe that shorter cycles incur higher FLOPs, while excessively long cycles result in noticeable quality degradation. The setting $N=5$ provides a favorable trade-off, preserving generation quality while reducing computation. Similarly, for cache ratio $R$, lowering $R$ below $0.7$ significantly harms quality, whereas $R=0.9$ eliminates redundant computation without compromising fidelity. Therefore, we adopt $N=5$ and $R=0.9$ as default hyperparameters in our main experiments.

\section{Conclusion}
This paper first gives an in-depth study of the caching error on aggressive and conservative methods and then introduces dual caching \texttt{DuCa}, which performs aggressive and conservative alternatively to leverage their advantages in different caching steps. Moreover, we have discovered the value of random selection and further explored its implications, revealing that diversity plays a more critical role in token selection than previously thought. Random selection not only demonstrates excellent performance but also seamlessly integrates with efficient attention methods like FlashAttention, making feature caching a practical and viable approach in real-world applications. We hope this paper can attract more attention to feature caching for its real benefits, to make it simple but effective, instead of complex.

\bibliographystyle{IEEEtran}
\bibliography{main}

\end{document}